\documentclass[conference]{IEEEtran}
\IEEEoverridecommandlockouts
\usepackage{cite}
\usepackage{amsmath,amssymb,amsfonts,bm}
\usepackage{graphicx}
\usepackage{textcomp}
\usepackage{xcolor}
\usepackage{algorithm}
\usepackage{algpseudocode}
\usepackage{subcaption}
\usepackage{balance}

\makeatletter
\newcommand{\linebreakand}{%
  \end{@IEEEauthorhalign}
  \hfill\mbox{}\par
  \mbox{}\hfill\begin{@IEEEauthorhalign}
}
\makeatother

\def\BibTeX{{\rm B\kern-.05em{\sc i\kern-.025em b}\kern-.08em
    T\kern-.1667em\lower.7ex\hbox{E}\kern-.125emX}}
\begin{document}

\title{Evaluation of a Canonical Image Representation for Sidescan Sonar} 

\author{\IEEEauthorblockN{Weiqi Xu \textit{weiqi@kth.se}}
\IEEEauthorblockA{\textit{KTH - Royal Institute of Technology} \\
Stockholm, Sweden 
}
\and
\IEEEauthorblockN{Li Ling \textit{liling@kth.se}}
\IEEEauthorblockA{\textit{KTH - Royal Institute of Technology} \\
Stockholm, Sweden 
}\and
\IEEEauthorblockN{Yiping Xie \textit{yipingx@kth.se}}
\IEEEauthorblockA{\textit{KTH - Royal Institute of Technology} \\
Stockholm, Sweden 
}\linebreakand
\IEEEauthorblockN{Jun Zhang \textit{juzhang@kth.se}}
\IEEEauthorblockA{\textit{KTH - Royal Institute of Technology} \\
Stockholm, Sweden 
}\and
\IEEEauthorblockN{John Folkesson \textit{johnf@kth.se}}
\IEEEauthorblockA{\textit{KTH - Royal Institute of Technology} \\
Stockholm, Sweden 
}}

\maketitle

\begin{abstract}
Acoustic sensors play an important role in autonomous underwater vehicles
(AUVs). Sidescan sonar (SSS) detects a wide range and provides photo-realistic images in high resolution. However, SSS projects the 3D seafloor to 2D images, which are distorted by the AUV’s altitude, target’s range and sensor’s resolution. As a result, the same physical area can show significant visual differences in SSS images from different survey lines, causing difficulties in tasks such as pixel correspondence and template matching. 
In this paper,  a canonical transformation method consisting of intensity correction and slant range correction is proposed to decrease the above distortion. The intensity correction includes beam pattern correction and incident angle correction using three different Lambertian laws ($\cos$, $\cos^2$, $\cot$), whereas the slant range correction removes the nadir zone and projects the position of SSS elements into equally horizontally spaced, view-point independent bins.
The proposed method is evaluated on real data collected by a HUGIN AUV, with manually-annotated pixel correspondence as ground truth reference. Experimental results on patch pairs compare similarity measures and keypoint descriptor matching. The results show that the canonical transformation can improve the patch similarity, as well as SIFT descriptor matching accuracy in different images where the same physical area was ensonified.
\end{abstract}


\section{Introduction} \label{sec:intro}

Autonomous underwater vehicles (AUVs) equipped with sonars allow us to collect acoustic returns from  the seabed at closer distances, thus providing data with higher resolution compared to surface vessels. However, without GPS, the pose estimate of the AUVs is prone to accumulative drift.  
Accurate pose estimation is of paramount importance when exploring or inspecting underwater environments with AUVs, since the oceanographic data collected need to be accurately geo-referenced. Acoustic ranging systems such as ultra short baseline systems (USBL) are often pre-installed on-site to help reduce the AUV's position estimates in commercial surveys. However, such systems are costly and time-consuming to deploy, and significantly restrict the AUV's survey area.

\begin{figure}[hbt!]
    \centering
    \begin{subfigure}{0.11\textwidth}
        \includegraphics[width=\textwidth]{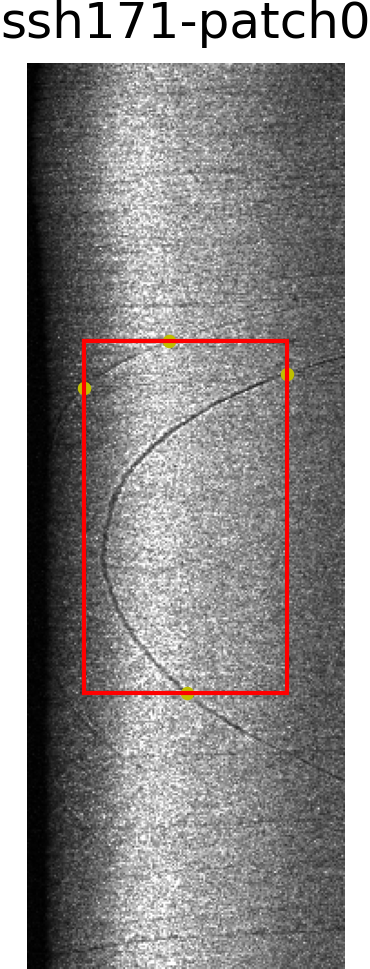}
    \end{subfigure}
    \begin{subfigure}{0.11\textwidth}
        \includegraphics[width=\textwidth]{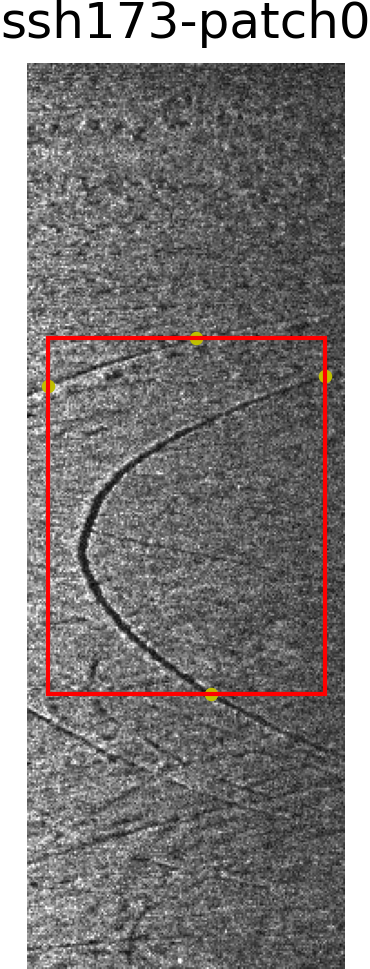}
    \end{subfigure}
    \begin{subfigure}{0.11\textwidth}
        \includegraphics[width=\textwidth]{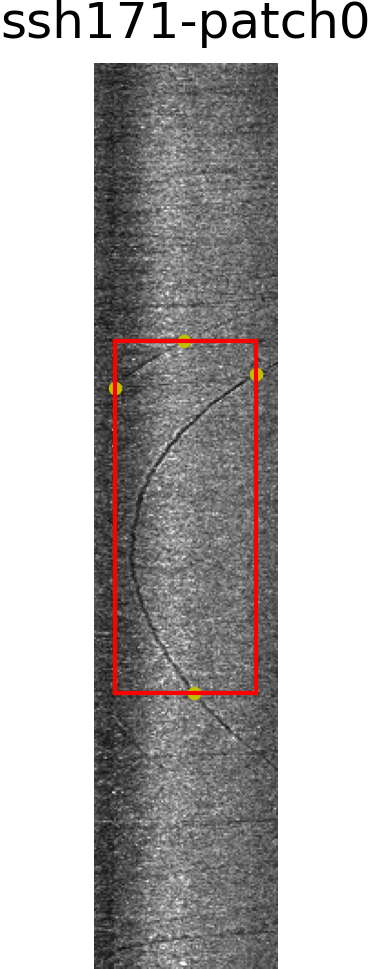}
    \end{subfigure}
    \begin{subfigure}{0.11\textwidth}
        \includegraphics[width=\textwidth]{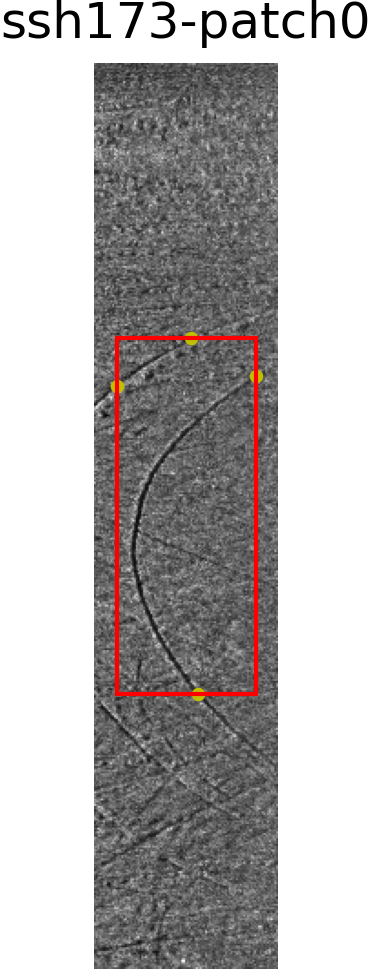}
    \end{subfigure}
  	\caption{The same physical seafloor area in SSS patches extracted from different survey lines. The left two patches are extracted from raw images, and the right two patches are extracted from canonical images generated using th $\cos^2$ Lambertian law. The red rectangle highlights the same physical pattern across all images.}
 	\label{fig:size_compare}
\end{figure}

Simultaneous localization and mapping (SLAM)~\cite{thrun2002acm} offers a more flexible and cost-effective way of reducing drift in vehicle pose estimates. In underwater SLAM, the most frequently used sensors include underwater camera, multibeam echosounder (MBES), and sidescan sonar (SSS) \cite{koser2020challenges, torroba2019towards, siantidis2016side}. Compared to cameras, SSS allows for sensing of much longer range. Compared to MBES, SSS can often have a higher resolution and is lower cost and easier to mount on an AUV.
Using SSS images for pose correction in any SLAM framework requires successful loop closure detection, or the detection of overlap in SSS images collected from different survey lines.
However, detecting matching feature pairs in SSS for loop closures is challenging, since the features on the seabed will appear differently in the SSS images collected from different distances and angles. 
The pixel intensity (due to  reflection and transmission) and resolution (due to geometric distortions) of SSS image are affected by multiple factors, including the vehicle's altitude, velocity, the sensor properties as well as the physical properties of the ensonified seafloor \cite{Blondel2009, burguera2014intensity}. As a result, SSS images of the same physical area will differ from one another, necessitating intensity and range correction prior to the feature correspondence matching tasks. In particular, an image patch will be distorted compared to an orthographic projection of the ensonified region and will vary in intensity as a function of the sonar position, see Fig.~\ref{fig:size_compare} (left).  


In this paper, we present a method capable of transforming SSS images from different survey lines into a canonical representation where the affect of the aforementioned SSS distortions is reduced. The proposed method extends previously proposed SSS normalization algorithms in two ways: 1) Intensity corrections using three different relationships with incidence angles are compared. 2) The proposed algorithm allows for an arbitrary size of pixel resolution, making it easier to compare SSS images across sensors and settings. Such a canonical representation of SSS images allows us to create datasets from different surveys, which is essential for learning-based approach to tackle the data association problem in SSS SLAM.

To validate the effectiveness of our proposed canonical transformation, SSS images of the \textit{same} physical area taken from multiple survey lines are used. Compared to the raw SSS images, we show experimentally that SSS images in the canonical representation obtain significantly higher image similarity scores and better descriptor matching accuracy. In summary, the contributions of this work are:
\begin{enumerate}[\IEEEsetlabelwidth{3)}]
\item Presents a canonical transformation method with detailed comparison of three Lambertian laws, and a sensor independent slant range correction algorithm which allows arbitrary ground range resolution.
\item Conducts extensive experiments on real data with annotated keypoints as ground truth reference, using metrics of correlation measure, histogram distance and descriptor matching, to prove the effectiveness of canonical representation of SSS image.
\item Demonstrates results of improvements in canonical images over raw images in patch similarity and descriptor matching accuracy, showing great potentials in application to downstream tasks such as SSS image mosaicking and SLAM, etc.
\end{enumerate}
%

\section{Related Work}
\label{sec:related_work}
Sidescan sonar images are affected by both geometric and intensity distortions. Geometric distortions can be addressed by slant range correction and corrections with respect to sonar's altitudes and AUV's speed. Intensity distortion can be corrected by accounting for sonar's beam pattern, incidence angle variations and seabed sediments variations. 

One of early work is Cobra et al.~\cite{cobra1992geometric}, where they only use the sonar images themselves to estimate and correct geometric distortions introduced by motion instabilities of the towfish, whose stability is much worse than AUVs. 
Chavez~\textit{et~al.}~\cite{chavez2002processing} give a overview on how to apply both geometric distortions and intensity distortions to create a mosaic from multiple sidescan images.
Burguera and Oliver~\cite{burguera2014intensity, burguera2016high} propose a detailed SSS correction model under the flat sea floor assumption. The model includes intensity correction procedure with a physical ensonification intensity model to compensate the sonar's beam pattern, and a slant range correction to remove geometric distortions. The slant range correction is sensor dependant, since the ground range resolution is chosen to be the same as the slant range resolution, making the algorithm a simple linear interpolation between two corresponding adjacent bins. 
Zhao~\textit{et~al.}~\cite{zhao2017new} propose a new intensity correction method addressing the variations of the seabed sediment with an unsupervised sediment classification algorithm.

The time-varying gain (TVG) is usually applied to the sidescan returned echos onboard during the mission to compensate the intensity variations due to the range effect. However, the exact parameters of the TVG are usually unavailable. Capus~\textit{et~al.}~\cite{capus2004compensation} propose a method to estimate the range-dependent residual TVG effects together with the angle-dependent beam pattern, which reduces some artefacts in sidescan images.

The distortion estimation and correction are usually the pre-processing steps for specific applications such as mosaicking~\cite{reed2006fusion}, mapping~\cite{burguera2016high}, localization~\cite{peng2021improved} and SLAM~\cite{zhang2023arxiv}. Burguera and Bonin-Font applied the correction methods in~\cite{burguera2016high} to remove the artifacts and distortions in the SSS images before feeding them to a CNN for segmentation~\cite{burguera2020line}. In this paper, we demonstrate the effectiveness of canonical transformation on reducing geometric and intensity distortions. The proposed method can be used as a pre-processing step for feature matching~\cite{vandrish2011side,king2013oceans} and loop closure detection~\cite{larsson2020latent} in SLAM.

\section{Canonical transformation of SSS} \label{sec:method}
Due to the variability and distortions of SSS images mentioned in Section \ref{sec:intro},  works that perform automation on SSS images normally include a pre-processing step. In general, this pre-processing consists of intensity correction and slant range correction \cite{coiras2007multiresolution, Blondel2009, burguera2014intensity, burguera2016high}, and our proposed method follows the same schema.

\subsubsection{Intensity Correction}
Intensity correction aims to normalize the intensities across the image. The goal is to have a perfectly flat/horizontal seafloor producing a perfectly uniform intensity. The actual image will then only show differences due to deviations from that assumption. To perform such correction, the SSS backscatter intensity is usually modelled using the \textit{Lambertian} model \cite{coiras2007multiresolution, burguera2014intensity}:
\begin{equation} \label{eq:lambertian-model}
I(p)=K \Phi(p) R(p)|\cos (\theta(p))|,
\end{equation}
where $K$ is the normalization constant, $p$ is the reflection point, and $\Phi(p), R(p), \theta(p)$ are the beam pattern, reflectivity and incident angle at point $p$, respectively. We model the beam pattern as the theoretical model of a linear array with a large number of elements, similar as in~\cite{xie2022towards}:
\begin{equation}
    \Phi(\phi(p))=\bigg(\frac{k_{\phi}\cdot sin(\phi-\phi_0)}{sin(k_{\phi}\cdot sin(\phi-\phi_0))}\bigg)^4,
\end{equation}
where $k_{\phi}=2.78$ for a 3 dB beam width of $60^\circ$ and $\phi_0=30^\circ$ is the angle of depression that the sonar is mounted. The value of both beam width and $\phi_0$ are sensor dependent.

An intensity $\Tilde{I}$ free from incidence angle is thereby achieved using Equation \ref{eq:intensity-correction}:
\begin{equation} \label{eq:intensity-correction}
 \Tilde{I}(p)=\frac{I(p)\sec (\theta(p))}{\Phi(\phi(p))}.
\end{equation}
The $\sec(\theta(p))$ implies that the backscatter intensity is proportional to the cosine of the incidence angle. However, cosine square ($\cos^2$) \cite{aykin2013forward, thorsos2001overview} and cotangent ($\cot$) \cite{folkesson2020lambert} relationships have also been suggested. In this work, we explore all three relationships for the intensity correction module and compare the resulting canonical representations using similarity scores and descriptor matching accuracies. The evaluation results are found in Section \ref{sec:exp}.

\subsubsection{Sensor Independent Slant Range Correction}
\begin{figure}[tbh!]
	\includegraphics[width=0.5\textwidth]{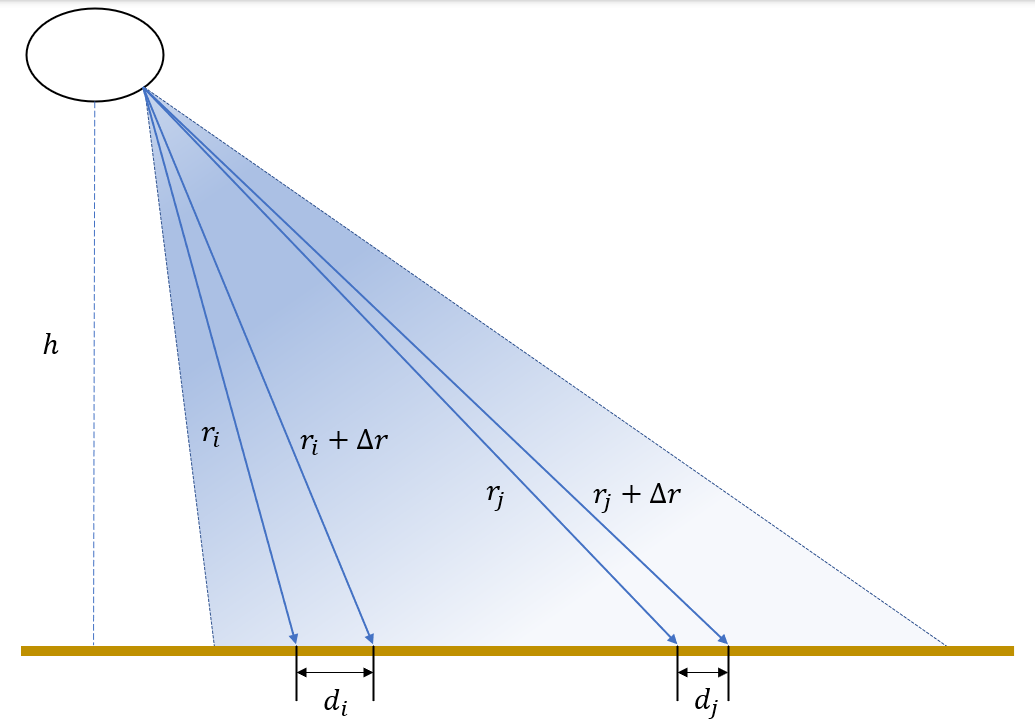}
	\caption{An intuitive comparison of ground range bin sizes at different slant ranges. With the same slant range resolution $\Delta r$, the effective ground range resolution increases as we move away from the sonar head.}
	\label{fig:slant2ground}
\end{figure}

This correction adjusts for the varying projection of the sidescan pixels on the (assumed) horizontal seafloor. The problem of geometric distortion in raw SSS images is visualized in Fig.~\ref{fig:slant2ground}. Although each consecutive bin/pixel is sampled at the same slant range resolution $\Delta r$, the resulting ground range resolution of $d_i$ and $d_j$ are different. As evident in Fig.~\ref{fig:slant2ground}, bins sampled with smaller incidence angle $\theta$ (i.e. closer to the sonar head) will cover larger physical area compared to bins sampled with larger incidence angle.

%

In our method, we first project all the sonar intensities to the assumed horizontal seafloor. Then we average and interpolate them to fixed horizontal size pixels. The interpolated intensities can be computed as the weighted sum of original intensities:
\begin{equation} \label{eq:3.9}
    \Bar{I_i} = \sum_j w_{ij} \Tilde{I_j},
\end{equation}
where $\Bar{I_i}$ is the $i$th bin in swath after canonical transformation, $\Tilde{I_j}$ is the $j$th bin after intensity correction, and $w_{ij}$ is the weight contributed to bin $\Bar{I_i}$ by $\Tilde{I_j}$, which is proportional to the corresponding distance of the $j$th bin along the horizontal line covered by the $i$th bin after interpolation. Here $w_{ij}$ is normalized so that $\sum_j w_{ij} = 1$. 

\begin{figure}[tbh!]
	\includegraphics[width=0.5\textwidth]{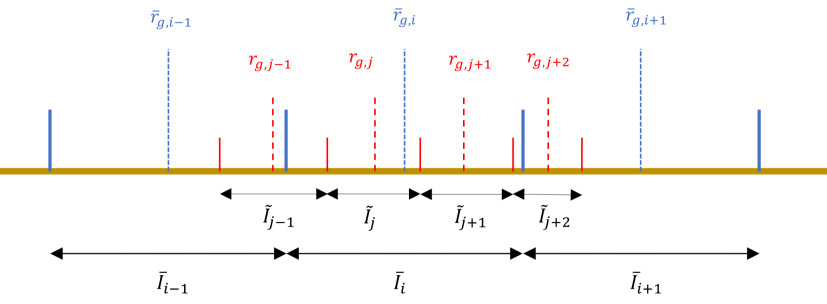}
	\caption{Interpolation in uniform ground range. The red solid vertical lines are the original points where the sidescan bin range hits the horizontal seafloor and the red dashed lines are the midpoints. The blue solid vertical lines are the evenly spaced ground range points and the blue dashed vertical lines are the midpoints \cite{weiqi22}.}
	\label{fig:rangeCorrect}
\end{figure}

A concrete schema is shown in Fig.~\ref{fig:rangeCorrect}, where $r_{g,j}$ and $\bar{r}_{g,i}$ are the ground range before and after uniformly spacing, respectively. The fixed size $\bar{I}_i$ is the weighted sum of $\Tilde{I}_{j-1},\Tilde{I}_{j},\Tilde{I}_{j+1},\Tilde{I}_{j+2}$:
\begin{equation} \label{eq:intensity-interpolation}
\begin{split}
    \Bar{I}_i = &[(\frac{r_{g,j-1} + r_{g, j}}{2}) - (\Bar{r}_{g,i}-\frac{\delta_g}{2})] \Tilde{I}_{j-1}\\
    &+ (\frac{r_{g,j+1} - r_{g,j-1}}{2}) \Tilde{I}_j
     + (\frac{r_{g,j+2} - r_{g,j}}{2}) \Tilde{I}_{j+1} \\
    &+ [(\Bar{r}_{g,i}+\frac{\delta_g}{2}) - (\frac{r_{g,j+2} + r_{g, j+1}}{2})] \Tilde{I}_{j+2}.
\end{split}
\end{equation}

The pseudo code of the slant correction algorithm for one ping is given in Algorithm \ref{alg:1}. 
\begin{algorithm}
\caption{Slant range correction for one ping}\label{alg:1}
\begin{algorithmic}
\State \textbf{Input:} Intensity corrected ping $\Tilde{I}$ of length $m$
\ForAll{$i = 1, \dots, m$}
    \State $j_1 = \underset{j}{\arg \max} [(\Bar{r}_{g,i} - \frac{d}{2}) \in N(j)]$
    \State $j_n = \underset{j}{\arg \min} [(\Bar{r}_{g,i} + \frac{d}{2}) \in N(j)]$
    \State $w_{ij_0} = \frac{r_{g,j_0} + r_{g, j_0+1}}{2} - (\Bar{r}_{g,i}-\frac{\delta_g}{2})$
    \State $w_{ij_n} = (\Bar{r}_{g,i}+\frac{\delta_g}{2}) - \frac{r_{g,j_n} + r_{g, j_n-1}}{2}$
    \ForAll{$j = j_0+1, \dots, j_n-1$}
        \State $w_{ij} = \frac{r_{g,j+1} - r_{g,j-1}}{2} $
    \EndFor
    \State Normalize $[w_{ij_1}, \dots, w_{ij_n}]$
    \State $\Bar{I_i} = \sum_j w_{ij} \Tilde{I_j}$
\EndFor
\end{algorithmic}
\end{algorithm}

Given the altitude $h$ and maximum slant range $r_{s,\textrm{max}}$ of one SSS ping, we project all the sonar elements to the assumed horizontal seafloor starting from $\theta_0=30^\circ$ incidence angle, which gives us the ground range from $r_{g,\textrm{min}}=h\tan(\theta_0)$ to $r_{g,\textrm{max}}=\sqrt{r_{s,\textrm{max}}^2-h^2}$. Selected arbitrary ground range resolution $\delta_g$, we can calculate the intensity for each ground range interval as a weighted sum of the corresponding bins.

Note that by selecting $\theta_0=30^\circ$, the removal of the nadir zone backscatter intensities, or the intensities in regions where no seafloor is detected, is performed implicitly for our data, see example in Fig. \ref{fig:intutive_comparision}. The choice of $\theta_0$ is dependent on the sensor as well as the mounting angle.

%

\section{Experimental Results} \label{sec:exp}
\subsection{Data Preparation}
The dataset we use in this project was collected in a survey of a fjord by a Kongsberg Hugin AUV equipped with EM2040 Multibeam and Edgetech 2205 sidescan sonars. By conducting the $\emph{sidescan draping}$ process described in \cite{bore2020modeling}, bathymetry built from MBES data can associate the intensity for each bin to its georeferenced coordinates, which finds the potential pixel correspondences between SSS images automatically. Manual correction is later conducted to adjust the keypoint positions to register a correspondence.

\begin{figure}[tbh!]
    \begin{subfigure}{0.4\textwidth}
    	\includegraphics[width=\textwidth]{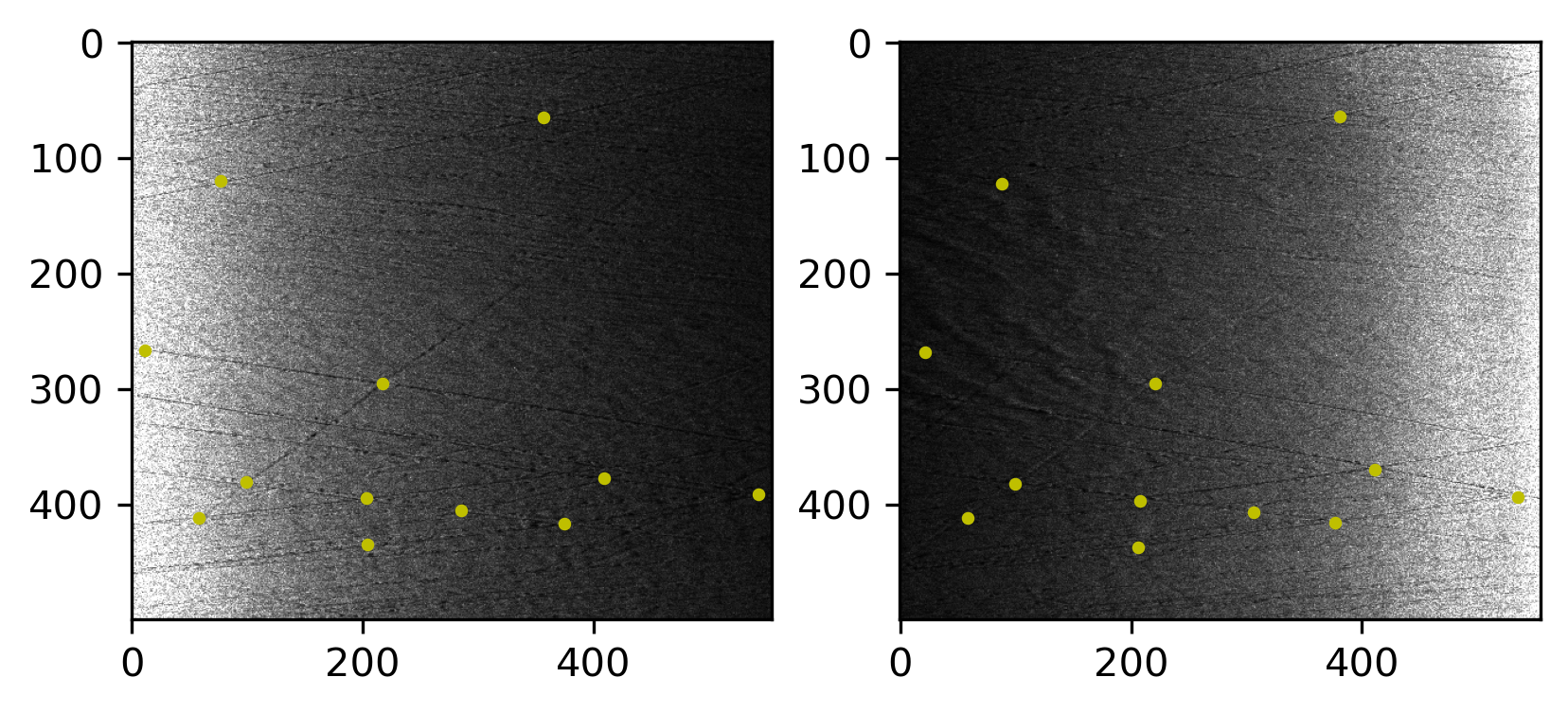}
    	\caption{Raw patch pair}
    	\label{fig:patchpair_raw}
    \end{subfigure}
    \begin{subfigure}{0.4\textwidth}
    	\includegraphics[width=\textwidth]{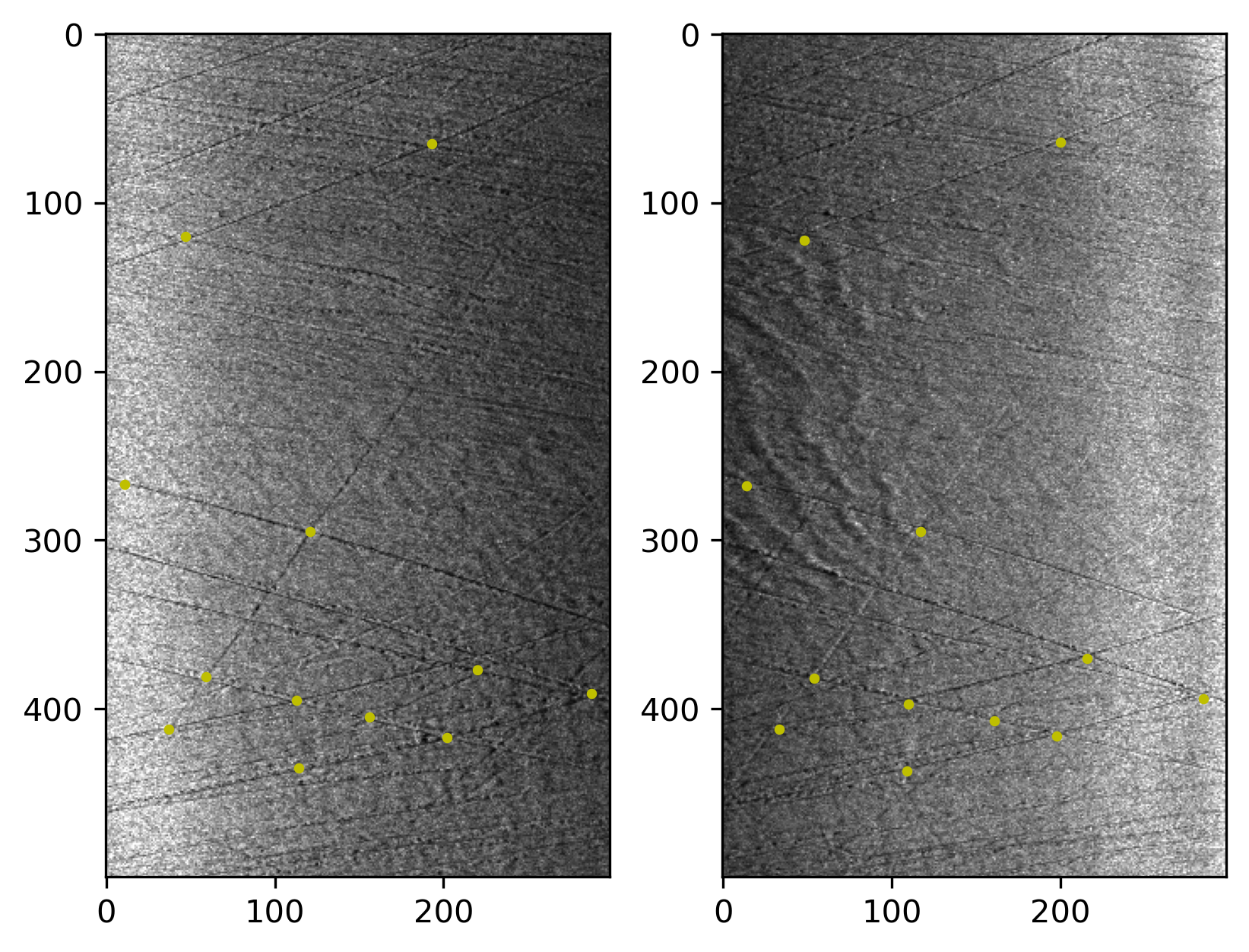}
    	\caption{Canonical patch pair}
    	\label{fig:patchpair_cano}
    \end{subfigure}
    \caption{Example patch pair extracted from two SSS images. (a) and (b) represents the same patches in raw and canonical images, respectively. The annotated keypoints are marked by the yellow dots.}
    \label{fig:patchpair}
\end{figure}

Given the annotated correspondences as ground truth reference, patch pairs are extracted from SSS images for the evaluation of the canonical transformation. The first step is to split the anchor waterfall image into starboard and port sides, since distance between pixels in two sides would not reflect real geometry after nadir zone removal. For each side, keypoint sets are selected in a constant size of ping set. In each pair of SSS images, patches are extracted around the corresponding keypoint sets in the same window sizes. An example of extracted patch pair in raw and canonical images is displayed in Fig. \ref{fig:patchpair}. Eventually a dataset with $1270$ keypoints correspondences along $13208$ pings and $60$ patch pairs was created.
%

\subsection{Evaluation Metrics}
To evaluate the effect of canonical transformation, two types of metrics are applied. One is intensity based methods including correlation measure and histogram distance between a patch pair to reflect the patch similarity. The other is descriptor based method evaluating the matching accuracy of keypoints between SSS images. For both metrics the baseline is computed using the patch pairs extracted from raw SSS images.

Correlation measure is usually applied to conduct pixel level comparison of two images of the same object taken in different time. Given $F(x,y)$ and $G(x,y)$ which represent the intensity in the same location $(x,y)$ of patches $F$ and $G$, the normalized correlation of the two patches is:
\begin{equation} \label{eq:3.15}
R(F,G)=\frac{\sum_{x, y}\left(F\left(x, y\right) \cdot G\left(x, y\right)\right)}{\sqrt{\sum_{x, y} F\left(x, y\right)^2 \cdot \sum_{x, y} G\left(x, y\right)^2}}.
\end{equation}
The range of correlation is in $[0,1]$, with a higher score indicating a better similarity between the evaluated patches. 

As pixel level offset exists during manual keypoint annotation, two histogram distance metrics, Kullback-Leibler divergence~\cite{csiszar1975divergence} and Chi-square distance~\cite{puzicha1997non}, are proposed as additional measures to improve robustness in patch similarity evaluation. Given the histograms of patch pairs $H_1$ and $H_2$ along the same bin range, the Kullback-Leibler divergence is:
\begin{equation} \label{eq:kl-d}
d_{KL}\left(H_1, H_2\right)=\sum_I H_1(I) \log \left(\frac{H_1(I)}{H_2(I)}\right),
\end{equation}
and the Chi-square distance is:
\begin{equation} \label{eq:chi}
d_{C}\left(H_1, H_2\right)=2\sum_I \left(\frac{(H_1(I)-H_2(I))^2}{H_1(I)+H_2(I)}\right).
\end{equation}
For both distances, a higher returned score represents increasing dissimilarity between the patch pair.

For descriptor based evaluation, we extract two representative hand-crafted descriptors, ORB~\cite{rublee2011orb} and SIFT~\cite{lowe2004distinctive}, at the annotated keypoints of each patch pair, and match them between the patch by comparing their descriptor distances. The accuracy of the found matches serves as metric.



%
\begin{figure*}[hbt!]
    \centering
\begin{subfigure}{0.26\textwidth}
    \includegraphics[width=\textwidth]{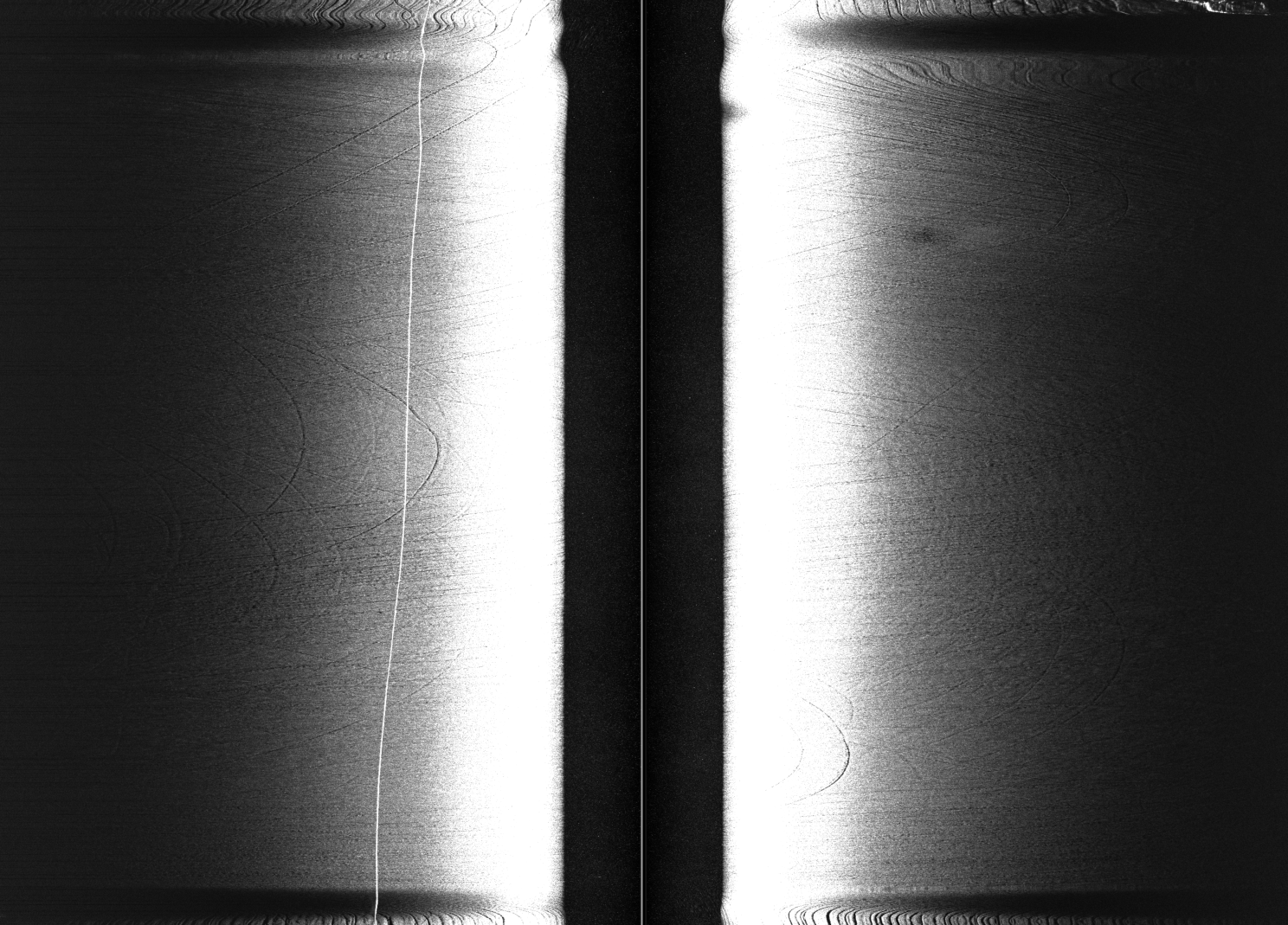}
    \caption{Raw}
	\label{fig:ssh170_0_raw}
\end{subfigure}
\begin{subfigure}{0.26\textwidth}
    \includegraphics[width=\textwidth]{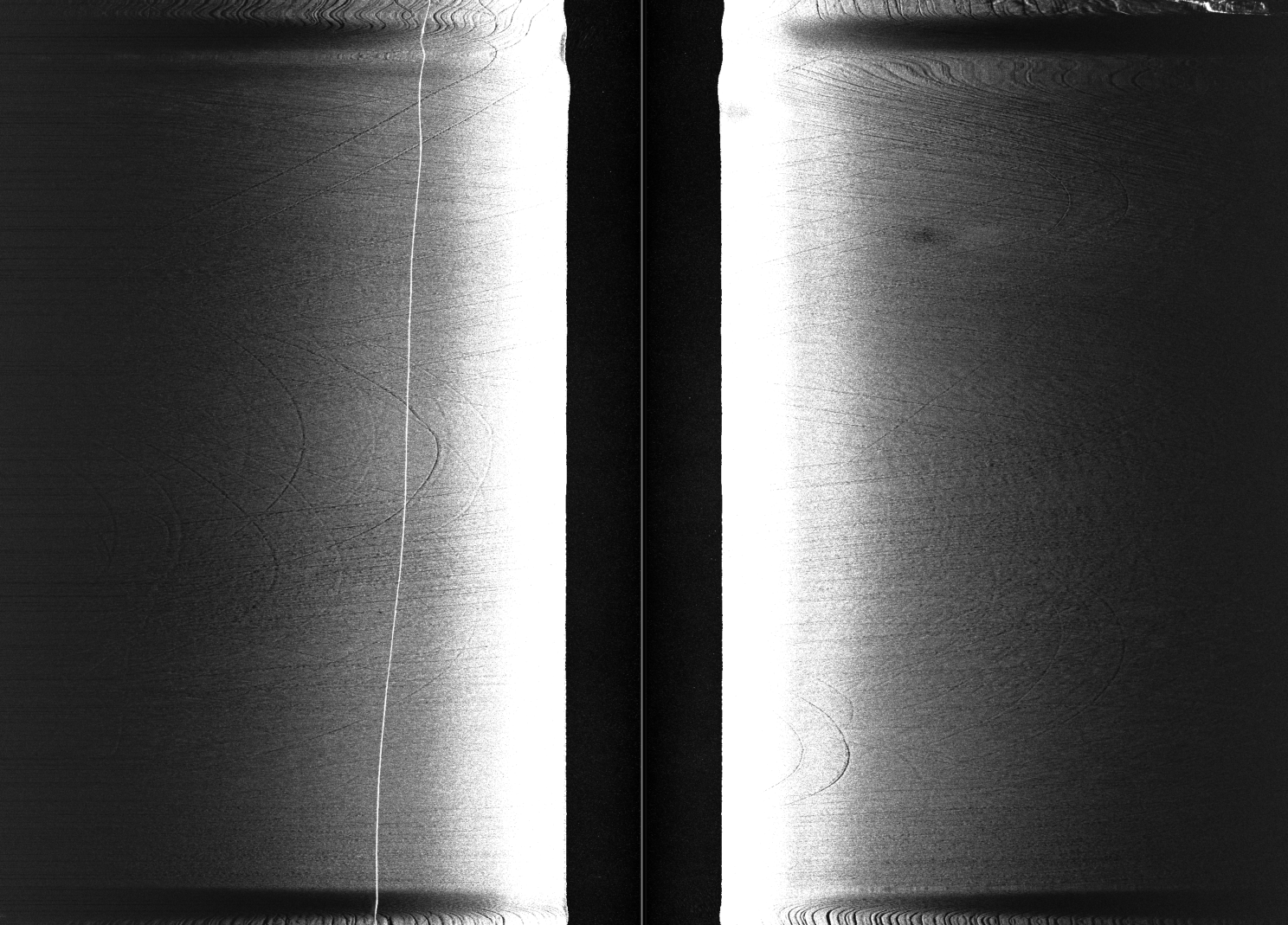}
    \caption{Beam pattern correction}
	\label{fig:ssh170_1_beam_pattern_correction}
\end{subfigure}
\begin{subfigure}{0.26\textwidth}
    \includegraphics[width=\textwidth]{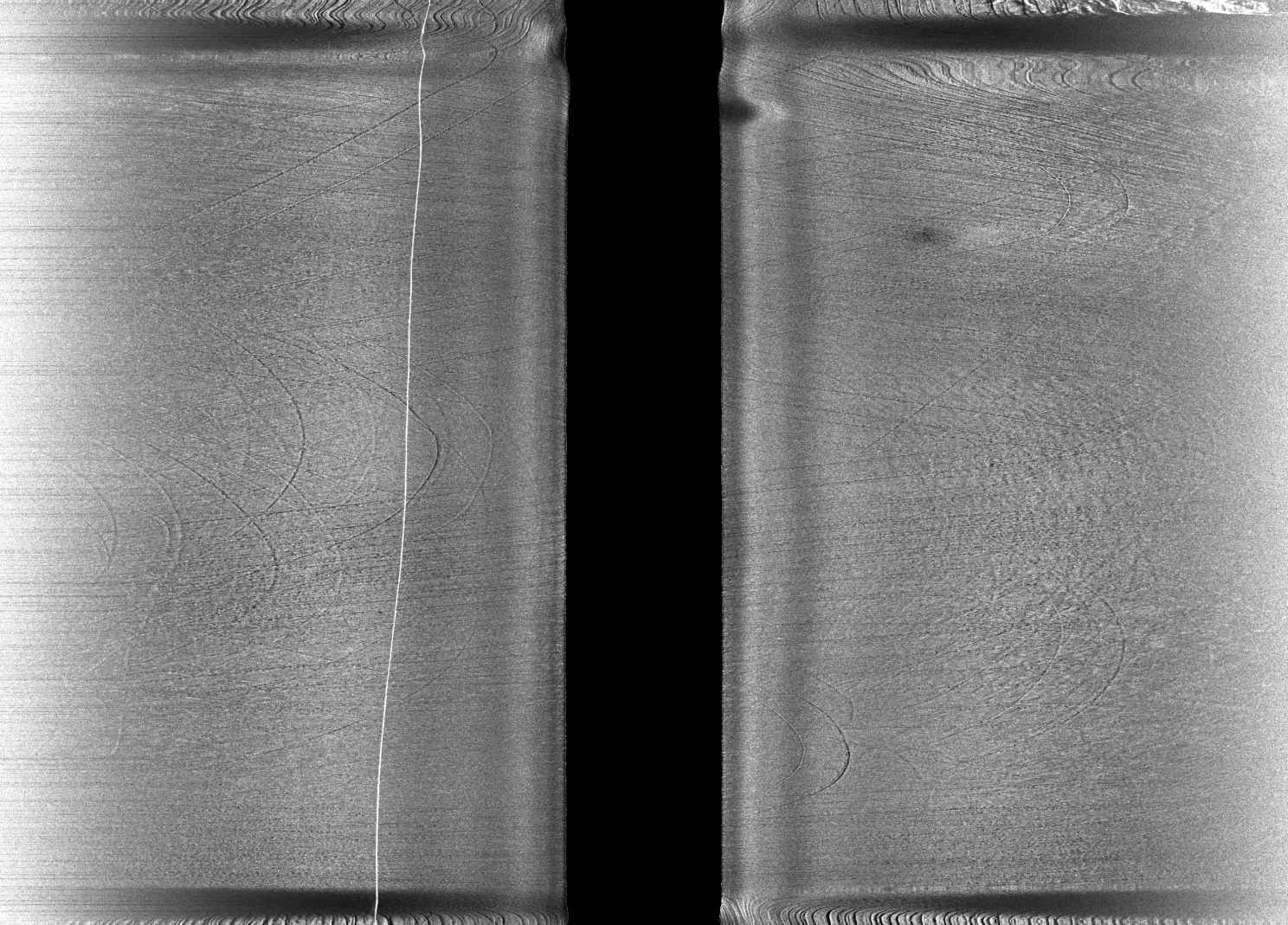}
    \caption{Incidence angle correction}
	\label{fig:ssh170_2_intensity_correction}
\end{subfigure}
\begin{subfigure}{0.18\textwidth}
\centering
    \includegraphics[width=0.68\textwidth]{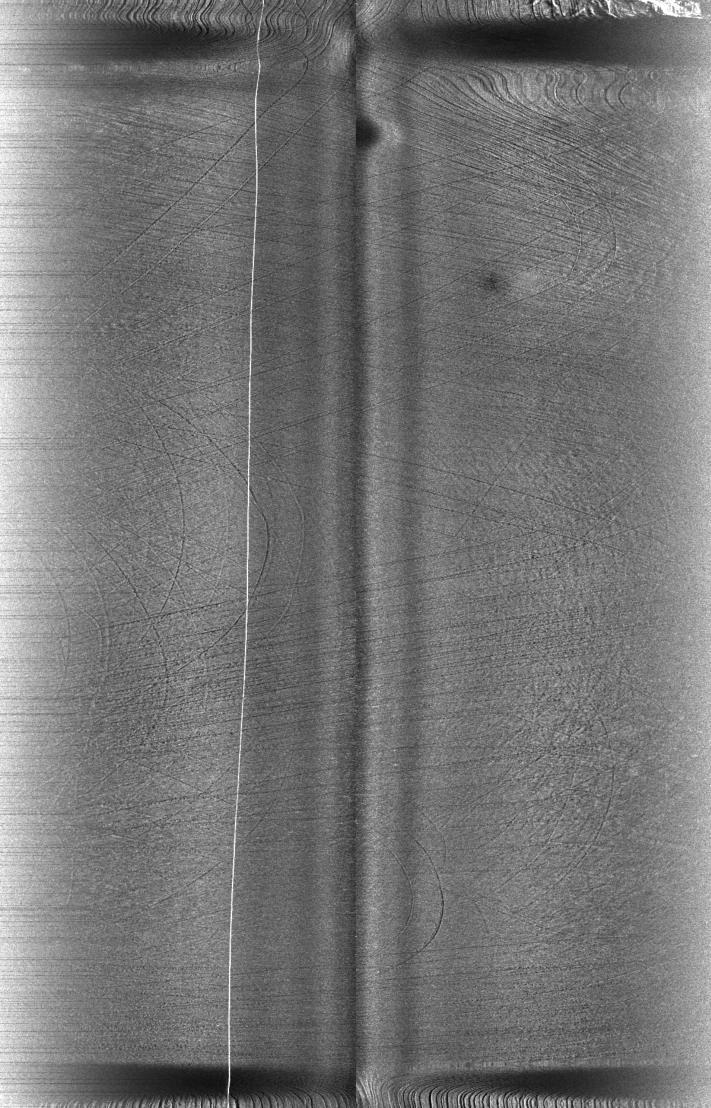}
    \caption{Slant range correction}
	\label{fig:ssh170_3_slant_range_correction}
\end{subfigure}
\caption{Comparison of a SSS waterfall image after each of the proposed transformation steps using $\cos^2$ law. (a)-(d) corresponds to the output after each step of the transformation, where (a) is the raw image, (b) is the beam pattern correction result, (c) is the incidence angle correction result and (d) is the slant range correction result, which is also the final output the proposed canonical transformation. The black strip in the center of (a)-(c) is the \textit{nadir zone} with low backscatter intensity and represents the distances where no seafloor is detected by the SSS. This region is removed in (d).}
\label{fig:intutive_comparision}
\end{figure*}
%

\subsection{Qualitative Results}
An intuitive comparison of each step in the canonical transformation using the $\cos^2$ Lambertian law is demonstrated in Fig.~\ref{fig:intutive_comparision}. In the raw image in Fig.~\ref{fig:ssh170_0_raw}, the center part of the image (regions closed to the nadir of the SSS) shows much stronger backscatter intensity compared to the far edges of the image. When visualizing the entire SSS image as a whole, this leads to information loss in both regions. Fig.~\ref{fig:ssh170_1_beam_pattern_correction} shows that beam pattern correction results in minor improvements. More significant improvement is seen in Fig.~\ref{fig:ssh170_2_intensity_correction} when intensity correction is applied. As the central areas with high echo intensities are normalized by reducing the $\cos^2$ Lambertian law, the intensities over the whole image look more even along swaths, which leads to a wider dynamic range for areas with useful information. In Fig.~\ref{fig:ssh170_2_intensity_correction}, a feature close to nadir in one image and far from the nadir in another image have more similar range of intensities. Fig.~\ref{fig:ssh170_3_slant_range_correction} shows the final result after slant range correction, where the nadir zone in the center is removed and the image appears to have more even scale.

Fig.~\ref{fig:size_compare} gives a closer inspection on effect of slant range correction on a feature's size.
In Fig.~\ref{fig:size_compare}, four keypoints are placed in the feature consisting of two curves, and a red rectangle is drawn to visualize the total area covered by these keypoints. In the raw patch pair, the window size is $112\times194$ and $153\times197$, respectively. The increasing width in raw patches demonstrates that the same feature covers more bins as it moves away from the nadir. In the canonical patches, the window size is $78\times194$ and $77\times197$, respectively. This indicates that the canonical transformation significantly improves the consistence of a feature's size in different ranges from the nadir by spacing bins uniformly along the ground range.
 
\subsection{Quantitative Results}

\subsubsection{Correlation Measure}
Here we compute the correlation similarity between the $60$ patch pairs in our dataset. For each canonical transformation method, the improvement in correlation is computed as $\rho= \frac{\Bar{R}-R}{R}$, with $R$ being the score on raw images and $\bar{R}$ the score on canonical images.
Table \ref{tab:correlation} shows the performance of raw and canonical patch pairs. All of the tested canonical transformations improved the correlation of $98.33\%$ of the patches, i.e. only 1 of the 60 patch pairs did not see improvement in correlation. Among them, the $\cos^2$ law achieved the highest performance, improving the average correlation from $0.6905$ by $34.50\%$ to $0.9287$.
%
\begin{table}[h]
    \centering
    \caption{Results in correlation measure for raw and canonical dataset. The highest values are highlighted in \textbf{bold}.}
    \label{tab:correlation}
\begin{tabular}{llll}
\hline Method 
       &\begin{tabular}[c]{@{}l@{}}Proportion of \\ Improved patches\end{tabular} 
       &\begin{tabular}[c]{@{}l@{}}Average \\ Similarity\end{tabular}  
       &\begin{tabular}[c]{@{}l@{}}Average \\ Improvement\end{tabular} \\

\hline 
None        & $-$       & $0.6905$      & $-$   \\
$\cos$      & $98.33\%$ & $0.8810$      & $27.59\%$    \\
$\cos ^2$   & $98.33\%$ & $\bm{0.9287}$ & $\bm{34.50}\%$    \\
$\cot$      & $98.33\%$ & $0.8988$      & $30.17\%$    \\
\hline
\end{tabular}
\end{table}
\subsubsection{Histogram Distance}
The result of histogram evaluations is listed in Table \ref{tab:Hist}. For each distance metric we analyse the proportion of improved patches after canonical transformation and average decrease $\rho = \frac{d-\Bar{d}}{d}$ using the average original distance $d$ and canonical distance $\Bar{d}$. The canonical transformation with $cos^2$ law again reaches the overall best performance in both metrics, with the average histogram distances in both metrics decreased by around $40\%$, and proportion of improved patches being $91.67\%$ and $83.33\%$, respectively.
\begin{table}[hbt!]
    \centering
    \caption{Results in histogram distance measure for raw and canonical dataset. The best values for each distance metric are highlighted in \textbf{bold}.}
\begin{tabular}{lllll}
\hline
  \begin{tabular}[c]{@{}l@{}}Distance \\ metric\end{tabular} &
  Method & 
  \begin{tabular}[c]{@{}l@{}}Proportion of \\ improved patches\end{tabular} &
  \begin{tabular}[c]{@{}l@{}}Average \\ score\end{tabular} &
  \begin{tabular}[c]{@{}l@{}}Average \\ decrease\end{tabular} \\ \hline


Chi-square & None    & -                & $485.37$        & -               \\
           & $\cos$   & $ 93.33\%$      & $339.72$        & $ 30.01\%$      \\
           & $\cos^2$ & $ 91.67\%$      & $\bm{298.59}$   & $\bm{38.48\%}$ \\
           & $\cot$   & $\bm{95.00}\%$  & $326.56$        & $ 32.72\%$      \\
\hline

KL-divergence & None    & -             & $1993.36$        & -               \\
           & $\cos$   & $ 70.00\%$      & $1586.12$        & $ 20.43\%$      \\
           & $\cos^2$ & $ \bm{83.33}\%$ & $\bm{1226.24}$   & $\bm{ 38.48\%}$ \\
           & $\cot$   & $ 75.00 \%$     & $1483.33$        & $ 25.59\%$      \\
\hline
\end{tabular}
\label{tab:Hist}
\end{table}

\subsubsection{Descriptor Matching}
Table~\ref{tab:descriptor} shows the descriptor matching results with the total number of proposed matches, the number of correct matches and the matching accuracy. 
For SIFT descriptors, all three tested canonical transformations lead to increase in number of total and correct matches, as well as higher matching accuracy. The best performance is again found using the $cos^2$ Lambertian law with a final matching accuracy of $92.68\%$, which is a slight improvement ($\approx 5\%)$ from the raw case. For ORB descriptors, the result is less conclusive. Out of the three canonical transformation cases, only $cos^2$ resulted in improved matching accuracy. It is also evident from Table~\ref{tab:descriptor} that out of the two tested descriptors, SIFT obtain both higher number of matches and higher accuracy and is more suitable for feature extraction from SSS images.

%
\begin{table}[h]
    \centering
    \caption{Matching results of ORB and SIFT descriptors for raw and canonical transformation. The best values for each distance metric are highlighted in \textbf{bold}.}
\begin{tabular}{lllll}
\hline Descriptors & Method & Total Matches & Correct Matches & Accuracy \\

\hline SIFT & None      &  $153$    & $133$     & $ 86.93 \%$ \\
            & $\cos$    & $160$     & $146$     & $  91.25\%$ \\
            & $\cos^2$  & $164$     & $\bm{152}$& $\bm{92.68}\%$ \\
            & $\cot$    & $168$     & $151$     & $  89.88\%$ \\
\hline ORB  & None     & $75$ & $50$        & $ 66.67\%$ \\
            & $\cos$   & $87$ & $\bm{55}$   & $  63.22\%$ \\
            & $\cos^2$ & $76$ & $54$      & $ \bm{71.05} \%$ \\
            & $\cot$   & $85$ & $\bm{55}$      & $  64.71\%$ \\
\hline
\end{tabular}
\label{tab:descriptor}
\end{table}

However, due to the limited number of image patches used for evaluation, definite conclusions on the effect of canonical transformation on descriptor matching performance cannot be drawn. Instead, we look for specific instances where canonical respective raw images achieve higher descriptor matching accuracy and analyse the qualitative differences between image patches. This analysis is only performed for SIFT descriptors due to its higher overall accuracy.

\begin{figure}[hbt!]
    \centering
    \includegraphics[width=.5\textwidth]{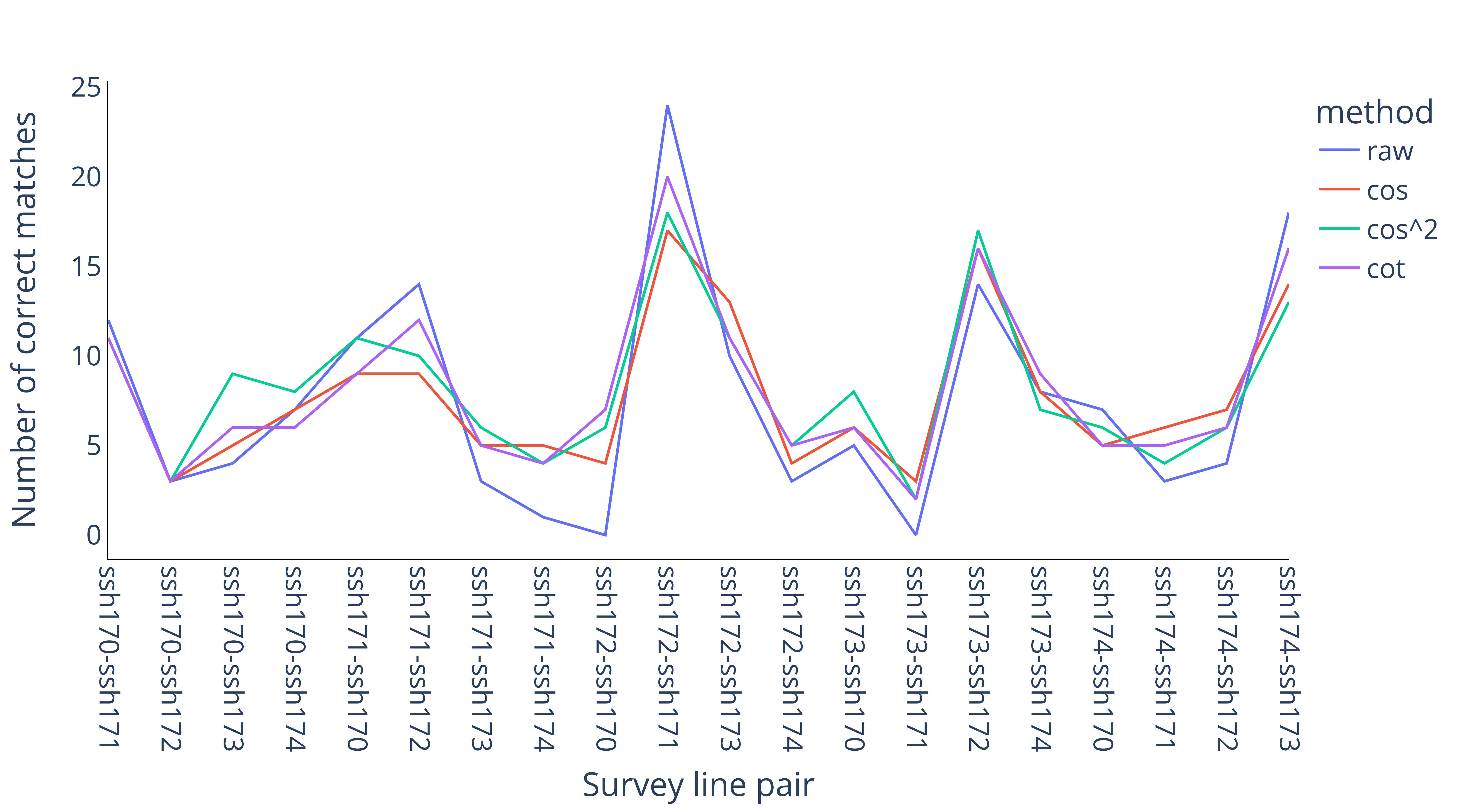}
    \caption{Correct number of matches found by using SIFT descriptor between all survey line pairs in the dataset.}
    \label{fig:sift-matches}
\end{figure}

\begin{figure}[hbt!]
    \centering
    \begin{subfigure}{.24\textwidth}
        \includegraphics[width=\textwidth]{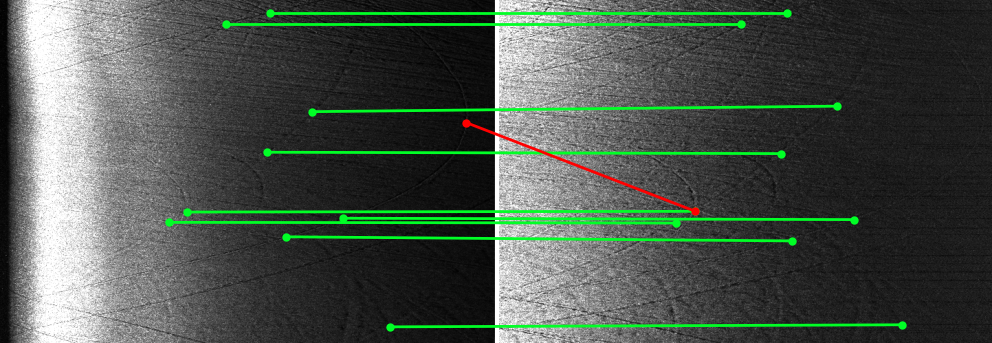}
        \caption{Raw}
        \label{fig:sift-raw-better-0-raw}
    \end{subfigure}
    \begin{subfigure}{.24\textwidth}
        \includegraphics[width=\textwidth]{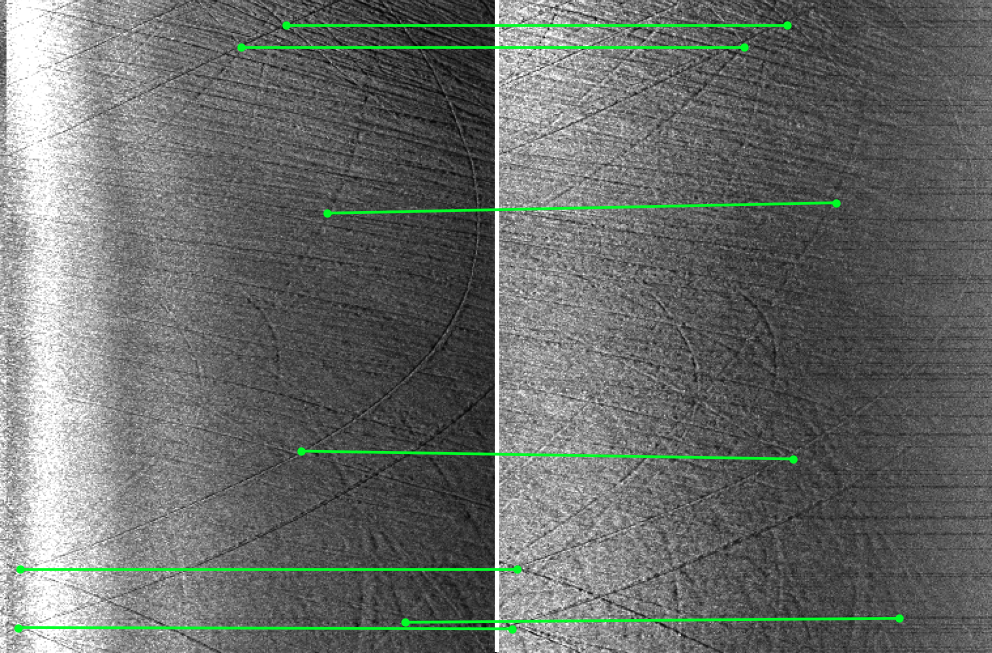}
        \caption{$cos$}
        \label{fig:sift-raw-better-1-cos}
    \end{subfigure}
    
    \begin{subfigure}{.24\textwidth}
        \includegraphics[width=\textwidth]{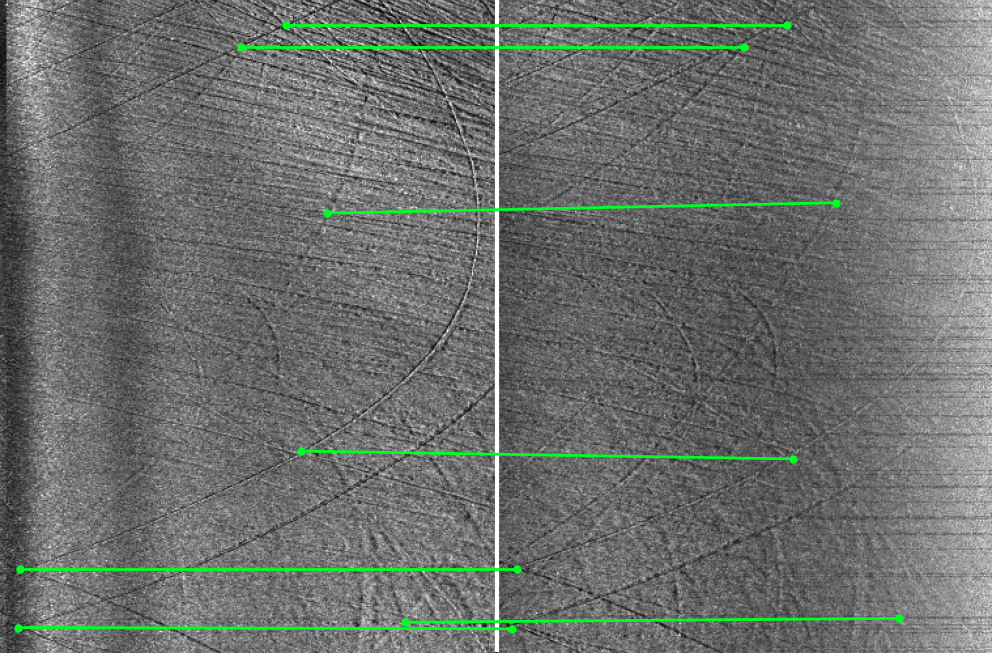}
        \caption{$cos^2$}
        \label{fig:sift-raw-better-2-cos^2}
    \end{subfigure}
    \begin{subfigure}{.24\textwidth}
        \includegraphics[width=\textwidth]{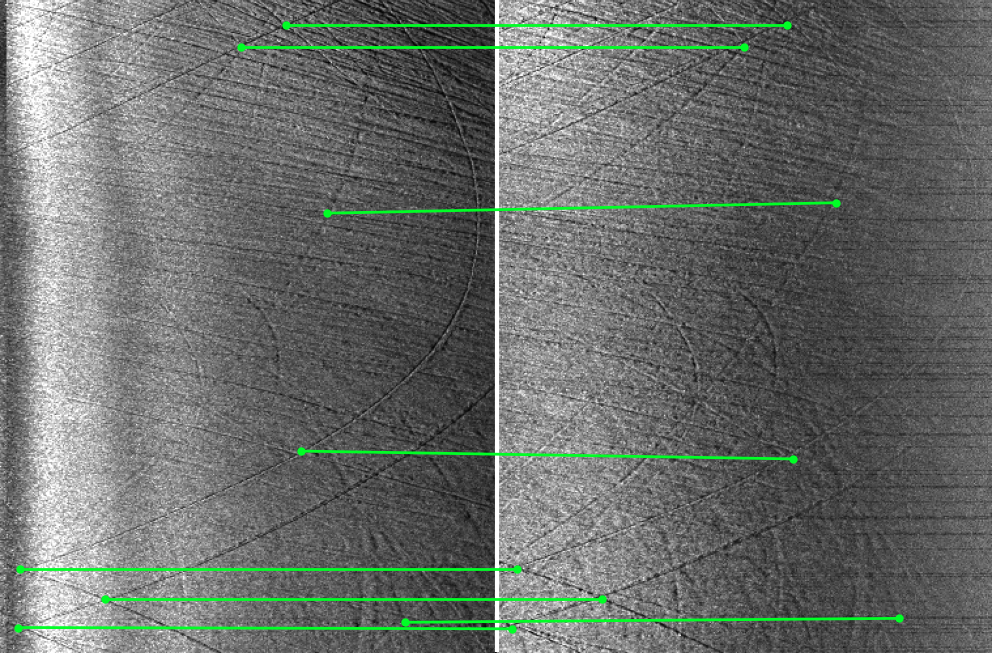}
        \caption{$cot$}
        \label{fig:sift-raw-better-3-cot}
    \end{subfigure}
    \caption{Example patch from the survey line pair \textit{ssh172-ssh171}, where the SIFT descriptor matching results on the raw image outperformed that on all three canonical images. The green and red lines indicate correctly and incorrectly proposed matches, respectively. In this example, the number of correct matches found in the raw image (a) is larger than in any other canonical images (b-d). One erroneous match is found in the raw image (a) and no erroneous matches are found in the canonical images (b-d).}
    \label{fig:sift-raw-better}
\end{figure}

\begin{figure}[hbt!]
    \centering
    \begin{subfigure}{.12\textwidth}
        \includegraphics[width=\textwidth]{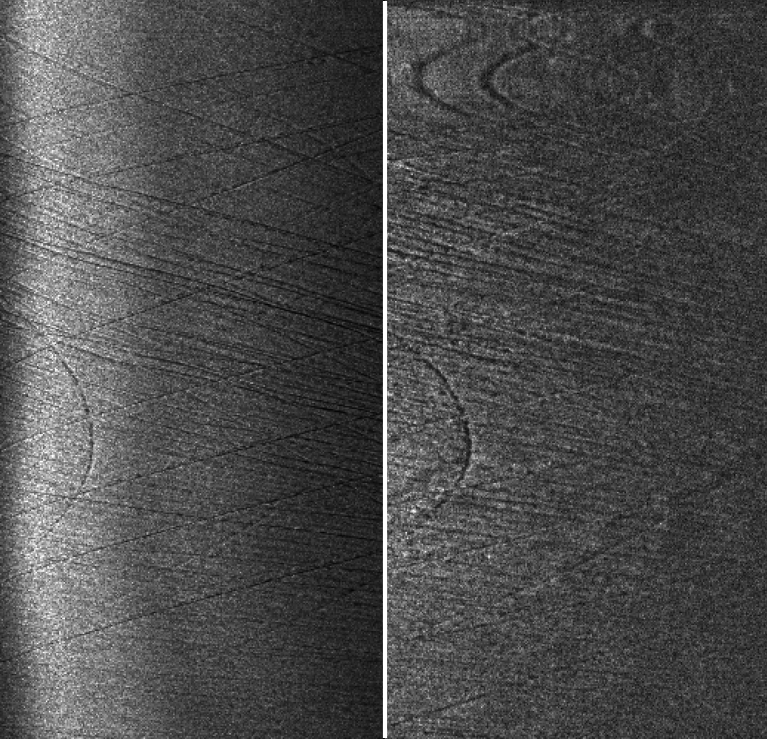}
        \caption{Raw}
        \label{fig:sift-cano-better-0-raw}
    \end{subfigure}
    \begin{subfigure}{.1\textwidth}
        \includegraphics[width=\textwidth]{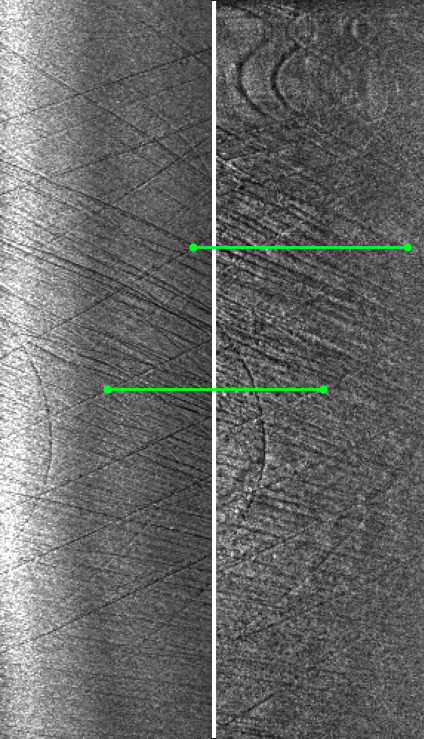}
        \caption{$cos$}
        \label{fig:sift-cano-better-1-cos}
    \end{subfigure}
    \begin{subfigure}{.1\textwidth}
        \includegraphics[width=\textwidth]{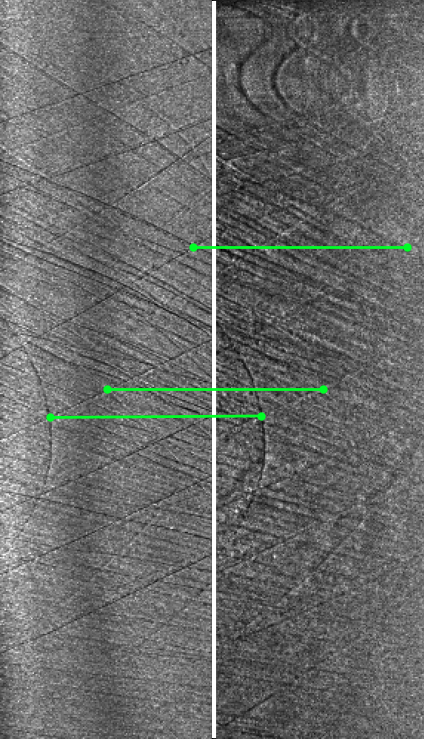}
        \caption{$cos^2$}
        \label{fig:sift-cano-better-2-cos^2}
    \end{subfigure}
    \begin{subfigure}{.1\textwidth}
        \includegraphics[width=\textwidth]{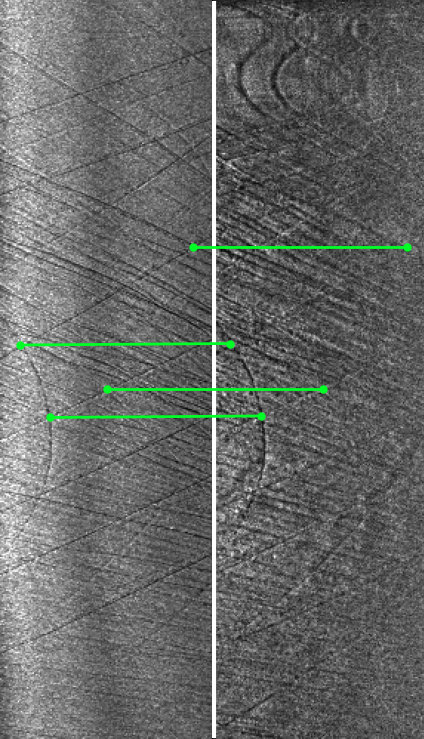}
        \caption{$cot$}
        \label{fig:sift-cano-better-cot}
    \end{subfigure}
    \caption{Example patch from the survey line pair \textit{ssh172-ssh170}, where the SIFT descriptor matching on all three canonical images outperformed that on the raw image. The green and red lines indicate correctly and incorrectly proposed matches, respectively. In this example, no matches is proposed in the raw image (a), and only correct matches have been proposed by the canonical images (b-d).}
    \label{fig:sift-cano-better}
\end{figure}

Fig.~\ref{fig:sift-matches} shows the correct number of matches proposed by SIFT across all patch pairs in the dataset, grouped by the survey line pairs where the patches are extracted from. The structure of the plot suggests that the performance of SIFT descriptors is indeed survey line dependent. For specific analysis, Fig.~\ref{fig:sift-raw-better} is an example from survey line pair \textit{ssh172-ssh171}, where the number of correct SIFT descriptor matches on the raw image outperforms that on all three canonical images. Fig.~\ref{fig:sift-cano-better} provides an opposite example from survey line pair \textit{ssh172-ssh170}, where SIFT obtained higher number of correct matches on all canonical images compared to on the raw image.

In Fig.~\ref{fig:sift-raw-better}, the two raw image patches both show significant but similar, intensity distortions compared to the canonical ones.  This might be why the canonical transformation does not help. 

In Fig.~\ref{fig:sift-cano-better}, the left image patch is extracted from closer to the nadir region, whilst the right image patch is extracted around the center. This makes the distortions in the raw image quite different in the two. Thus removing the distortions is more helpful.



From these examples, we hypothesise that the proposed canonical transformation can help preserve the repeatability of descriptors for keypoints lying in regions with large geometric and intensity distortions, though more extensive experiments are required for any conclusive results.

\subsubsection{Discussion}
The proposed canonical representation achieves great performance in intensity based metric, including correlation measure and two histogram distances. The results suggest that the $cos^2$ law achieves the best performance in increasing the similarity of patch pairs. 
For descriptor evaluations, the canonical transformation with $\cos^2$ Lambertian law can improve the matching accuracy for both SIFT and ORB descriptors by a small amount, as shown in Table~\ref{tab:descriptor}. However, the improvement is not consistent across all image pairs and more experiments should be conducted before any conclusive results can be drawn.


Since most significant improvement is found in patch similarity, our method is perhaps more suitable for tasks requiring canonical shape of patches for data associations, for instance, dense patch matching~\cite{barnes2009patchmatch,korman2015coherency}, which would be an interesting future direction. Furthermore, the proposed canonical transformation could be applied as a preprocessing step for other downstream tasks, such as place recognition using SSS images~\cite{larsson2020latent}, which is useful for loop closure detection in SSS SLAM.

\section{Conclusion}
\label{sec:concl}
In this paper we present a canonical transformation method for side-scan sonar. The proposed method is carefully evaluated on real data with annotated keypoints as ground truth, using two types of metrics. We have shown that canonical transformations of SSS image can lead to better patch similarity and descriptor matching accuracy, with the transformation using $\cos^2$ Lambertian law achieving the overall highest performance.

To guide future work, one could speculate that the proposed method could help enrich the data for training of neural networks by uniforming the SSS images from different locations (e.g, varying altitudes). The proposed method could also be helpful to other feature-based tasks such as object detection and sparse keypoint matching, as well as patch-based tasks such as loop closure detection for SSS SLAM, and dense patch matching for SSS image registration and mosaicking, etc.

\section*{Acknowledgment}

This work was partially supported by the Wallenberg AI, Autonomous Systems and Software Program (WASP) funded by the Knut and Alice Wallenberg Foundation and partially supported by Stiftelsen för Strategisk Forskning
(SSF) through the Swedish Maritime Robotics Centre (SMaRC)
(IRC15-0046).

\balance
\bibliographystyle{IEEEtran}
\bibliography{IEEEabrv,refs}

\end{document}